\documentclass{article}
\usepackage[T1]{fontenc}
\usepackage{graphicx} 
\usepackage{arxiv,times}
 
\usepackage{hyperref}
\usepackage{url}
\usepackage{graphicx}
\usepackage{booktabs}
\usepackage{multirow}
\usepackage{amsmath}
\usepackage{pifont}
\usepackage{xcolor}
\usepackage[most]{tcolorbox}
\usepackage{subcaption}

\title{ArtiVidRM: Verifiable Artifact-as-Rubric Reward Modeling for Explicit Quality Articulation in Video Generation}

\author{Caorui Li$^{1,2}$, Yangguang Ji$^{2}$, Shuxia Lin$^{1}$, Yu Chen$^{3}$, YiCheng Xiao$^{3}$, \\
\textbf{Yongliang Wu$^{2,4}$, Zhenxiang Jiang$^{2}$, Licheng Tang$^{2}$, Deng Li$^{2}$, Zhuxuanzi Wang$^{2}$,} \\
\textbf{Jiahao Sun$^{2}$, Chonghan Liu$^{5}$, Lirian Su$^{2}$, Jay Wu$^{2}$, Xu Yang$^{1}$} \\[4pt]
$^{1}$Southeast University \quad
$^{2}$Opus AI Research \\
$^{3}$Institute of Automation, Chinese Academy of Sciences \\
$^{4}$University of Illinois Urbana-Champaign \quad
$^{5}$University of California, Los Angeles
}

\arxfinalcopy
\begin{document}

\maketitle

\begin{abstract}
Reinforcement learning has become an effective approach for improving
video generation, but its success depends heavily on reliable reward
models. Existing methods either map videos directly to opaque scalar
scores or generate free-form rationales that are not explicitly optimized
as the basis of the final judgment.
We propose \textbf{ArtiVidRM}, a verifiable, artifact-as-rubrics video reward model that follows an observe-then-score paradigm. It first identifies visible generation artifacts and organizes them into a structured artifact list, then uses these artifacts as explicit rubrics for quality assessment. This intermediate representation grounds each score in concrete video-level evidence, making the reward more interpretable and informative for optimization.
We further introduce a \textbf{multi-reward reinforcement learning} strategy to jointly optimize artifact prediction and quality assessment. The reward decomposes into complementary objectives that evaluate artifact prediction quality, including semantic coverage and prediction efficiency, together with a pairwise preference objective that supervises quality ranking through score differences. By jointly optimizing the intermediate artifact representation and the downstream preference prediction, the model learns to produce informative artifact lists while improving the accuracy of video quality assessment.
Experiments show that ArtiVidRM achieves state-of-the-art
performance on video preference benchmarks, provides effective signals for
video diffusion optimization, and enables artifact-guided refinement of closed-source video generators.
\end{abstract}

\section{Introduction}
\label{sec:data_construction}

\begin{figure*}[t]
    \centering
    \includegraphics[width=\textwidth]{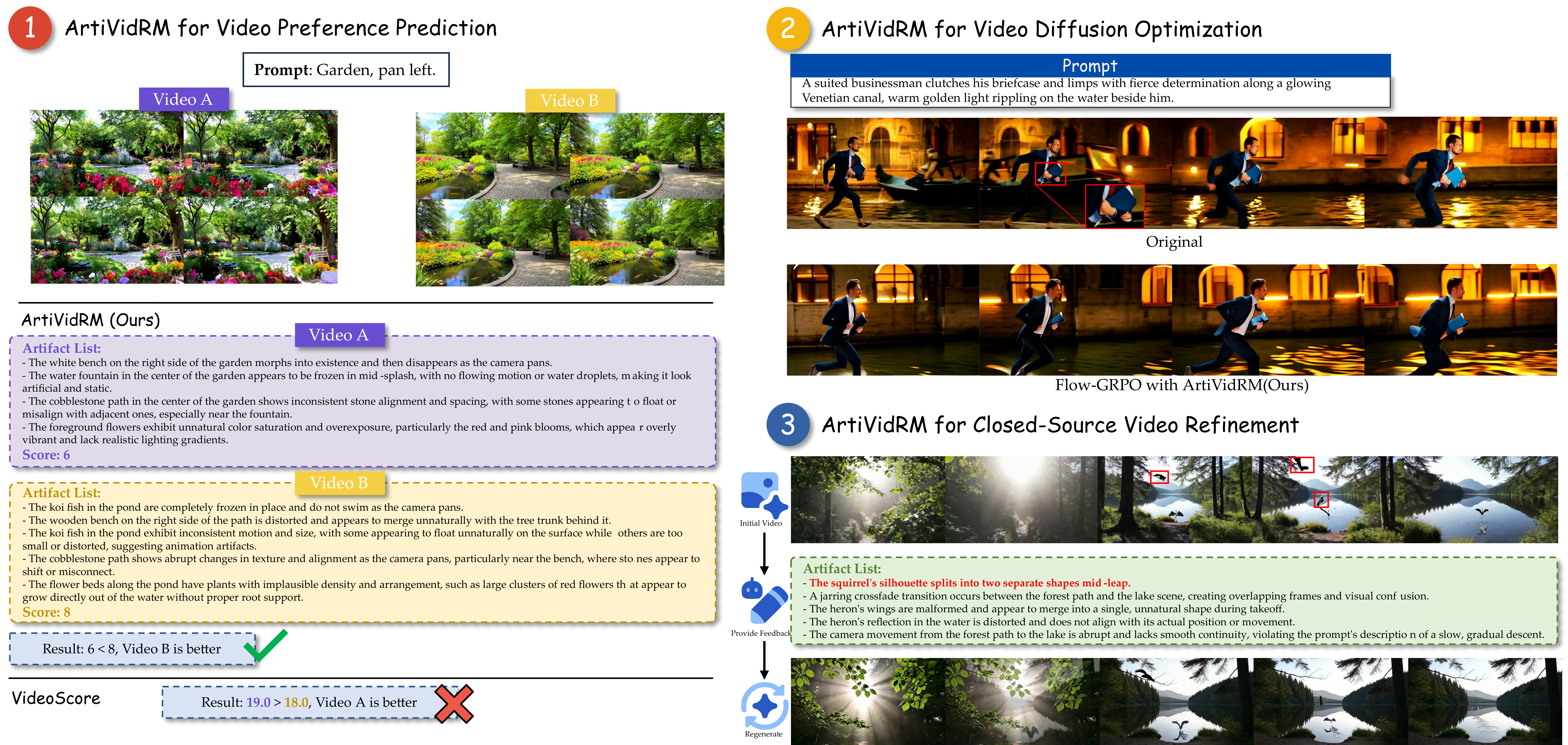}
    \caption{Overview of ArtiVidRM. By explicitly predicting structured artifact lists before quality scoring, ArtiVidRM enables (1) accurate video preference prediction, (2) reward-guided optimization of text-to-video diffusion models, and (3) artifact-guided refinement of closed-source video generators. The closed-source refinement example shown in the figure uses Veo 3.1.}
    \label{fig:intro}
\end{figure*}

Recent advances in video generation have led to significant overall improvements in the quality of generated videos~\cite{wan2025wanopenadvancedlargescale,lei2024comprehensivesurveyhumanvideo,ma2026controllablevideogenerationsurvey}. Reinforcement learning and preference optimization provide a promising way to further improve generation quality~\cite{shao2024deepseekmathpushinglimitsmathematical,deng2024flowdpoimprovingllmmathematical}, but their effectiveness depends critically on the reliability of video reward models. A useful reward model should not only determine whether a video is good or bad, but also reveal \emph{what is wrong} with it.

Existing approaches mainly fall into two categories. Scalar preference models directly map videos to numerical scores~\cite{he2024videoscorebuildingautomaticmetrics,liu2025improvingvideogenerationhuman}, but compress diverse artifacts into a single opaque value and provide little diagnostic feedback. Recent reasoning-based methods generate free-form rationales before scoring~\cite{he2025videoscore2thinkscoregenerative,wang2026unifiedrewardmodelmultimodal,xu2026visionrewardfinegrainedmultidimensionalhuman}, yet these rationales are not necessarily faithful to, or explicitly optimized as, the basis of the final judgment. Consequently, current reward models may predict human preferences without providing concrete and verifiable evidence for their decisions.

We propose \textbf{ArtiVidRM}, an artifact-as-rubric reward modeling framework for video generation, as illustrated in Figure~\ref{fig:intro}. It follows an \emph{observe-then-score} paradigm: the model first identifies visible artifacts and quality issues, organizes them into a structured artifact list, and then uses these artifacts as explicit rubrics for quality assessment. This intermediate representation grounds the final score in observable video-level evidence and provides actionable feedback for targeted generation refinement.

ArtiVidRM is trained in two stages. We first perform supervised fine-tuning on reference video-artifact annotations to establish structured artifact identification ability. We then apply reinforcement learning on paired human preference data to jointly optimize artifact discovery and preference prediction. artifact-level rewards encourage well-formed and semantically faithful artifact lists, while preference rewards align score differences with human judgments. Experiments show that ArtiVidRM achieves competitive performance on multiple video quality benchmarks, provides effective signals for video diffusion optimization, and enables targeted iterative refinement of closed-source video generators.

Our \textbf{contributions} are summarized as follows:
\begin{itemize}
    \item We introduce an artifact-as-rubric reward modeling framework that explicitly connects observable generation artifacts with video quality judgments.
    \item We develop a two-stage training strategy that jointly optimizes structured artifact identification and human preference alignment.
    \item We demonstrate its effectiveness in video preference prediction, video diffusion optimization, and closed-source iterative refinement.
\end{itemize}

\section{Related Work}
\paragraph{Text-to-Video Quality Assessment.}
Recent text-to-video models~\cite{liu2024sorareviewbackgroundtechnology,yang2025cogvideoxtexttovideodiffusionmodels,wan2025wanopenadvancedlargescale,seedance2026seedance20advancingvideo,wang2025klingfoleymultimodaldiffusiontransformer} have
substantially improved visual fidelity, temporal coherence, and prompt
following. Nevertheless, generated videos still exhibit spatial
distortions, motion inconsistencies, temporal flickering, and semantic
misalignment. Existing evaluation frameworks, such as
EvalCrafter~\cite{liu2024evalcrafterbenchmarkingevaluatinglarge}, VBench~\cite{huang2023vbenchcomprehensivebenchmarksuite}, and
T2VQA~\cite{kou2024subjectivealigneddatasetmetrictexttovideo}, assess video quality through predefined metrics or
learned quality predictors, while recent video reward models further
learn human preferences from pointwise scores or pairwise comparisons.

\paragraph{Reasoning-Based Video Reward Models.}
Early reward models such as VideoScore~\cite{he2024videoscorebuildingautomaticmetrics} directly predict multidimensional
quality scores or pairwise preferences, but provide limited diagnostic
evidence. As multimodal large language models have become capable of
fine-grained and temporally grounded video understanding~\cite{An_2026,li2026omnivideobenchaudiovisualunderstandingevaluation,chen2026omnireasonerthinkinglongaudiovideo,gao2026claimlevelrubricrewardsvideo},
recent generative reward models build on them to improve interpretability:
UnifiedReward~\cite{wang2026unifiedrewardmodelmultimodal} supports pointwise scoring and
pairwise ranking, VisionReward decomposes evaluation
into fine-grained binary questions, VideoScore2~\mbox{\cite{he2025videoscore2thinkscoregenerative}}
generates multidimensional scores with chain-of-thought reasoning, and
DeScore~\cite{wang2026thinkscoredecoupledreasoning} decouples reasoning generation from reward
prediction. ArtiVidRM differs by learning an open-vocabulary,
video-specific artifact list as an explicitly optimized intermediate
representation and directly using the predicted artifacts as rubrics
for preference assessment.

\paragraph{Preference Optimization for Video Generation.}
Learned reward models have increasingly been used to align image and
video generators through preference-based fine-tuning or reinforcement
learning~\cite{amini2024directpreferenceoptimizationoffset,zhu2024diffusionmodelsreinforcementlearning,chen2026omnireasonerthinkinglongaudiovideo,xiao2025pixclipachievingfinegrainedvisual}. Representative approaches include VideoDPO~\cite{liu2024videodpoomnipreferencealignmentvideo},
OnlineVPO~\cite{zhang2025onlinevpoalignvideodiffusion}, Flow-GRPO~\cite{liu2025flowgrpotrainingflowmatching}, and
DanceGRPO~\cite{xue2025dancegrpounleashinggrpovisual}. These studies mainly
investigate generator optimization under a given reward function,
whereas ArtiVidRM focuses on learning a reliable and interpretable
video reward model that supports preference prediction, reward-guided
training, and artifact-guided iterative refinement.

\section{Method}
\subsection{Artifact-Augmented Preference Data Construction}
\label{sec:data_construction}

\begin{figure*}[t]
    \centering
    \includegraphics[width=\textwidth]{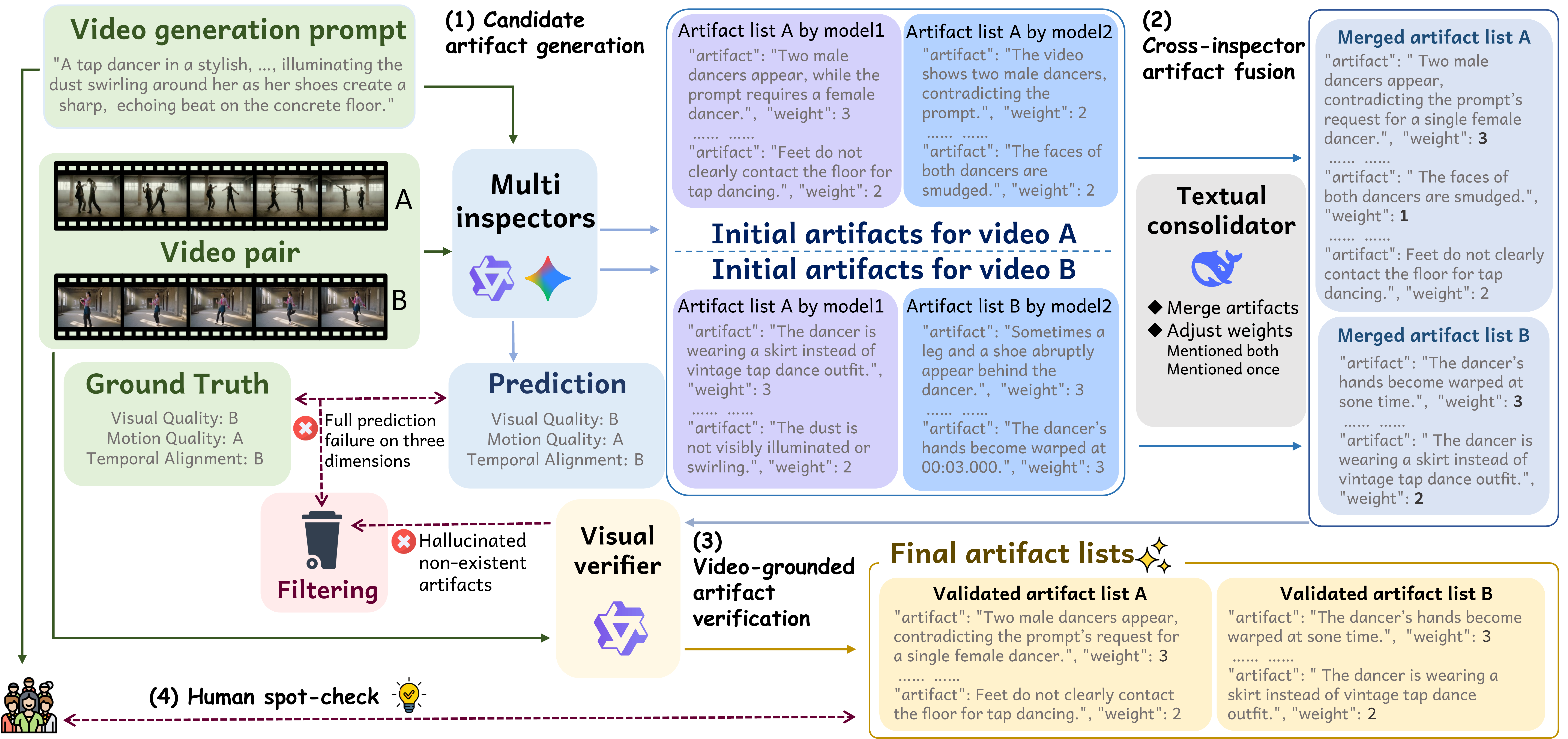}
    \caption{Overview of the artifact-augmented preference data construction pipeline.}
    \label{fig:framework}
\end{figure*}

\paragraph{Candidate artifact generation.}
We build on VideoGen-RewardBench~\cite{liu2025improvingvideogenerationhuman}, which contains 26.5K $(p, v^A, v^B)$ triplets with expert pairwise preference annotations, including an overall preference label and three aspect-specific labels: Visual Quality (VQ), Motion Quality (MQ), and Text Alignment (TA). For each video pair, we employ Gemini-2.5-Pro~\cite{comanici2025gemini25pushingfrontier} and Qwen3-VL-32B~\cite{yang2025qwen3technicalreport} as independent inspectors. Before identifying artifacts, each inspector first predicts the pairwise preferences along the three quality dimensions. A response is accepted only if at least two predictions agree with the human annotations; otherwise, the inspector is given one additional attempt. Responses that fail both attempts are discarded. Each candidate contains artifacts represented by a textual description and a severity score from 1 (minor) to 3 (severe).

\paragraph{Cross-inspector artifact fusion.}
The two inspectors may describe the same underlying artifact in different
wording or introduce inconsistent formats. We use DeepSeek-V4-Flash~\cite{deepseekai2026deepseekv4highlyefficientmilliontoken} as a
text-based consolidator to merge semantically equivalent entries,
normalize descriptions, and remove near-duplicates. Artifacts supported by only one inspector have their severity weight reduced by one to reflect lower confidence, while artifacts supported by both retain their higher weight.

\paragraph{Video-grounded artifact verification.}
Open-ended artifact generation is susceptible to hallucinations, whereas verifying whether a specific artifact is visible in a video is a more constrained and reliable task~\cite{bai2022constitutionalaiharmlessnessai,madaan2023selfrefineiterativerefinementselffeedback}.
We therefore introduce a video-grounded verification stage in which Qwen3-VL-32B, acting as a verifier, re-inspects each video against the consolidated artifact list and removes any entry that cannot be directly confirmed from visible evidence, removing 26.8\% of consolidated candidates, with a mean of 6.11 artifacts per video in the final candidate lists.

\paragraph{Final data format and human spot-check.}
Each processed instance contains the prompt, the verified artifact lists, the original human dimension-wise preference labels, and the overall pairwise preference label, all taken from the source dataset:
\begin{equation}
\mathcal{D}_i =
\left(
p_i,\,
v_i^A,\,
v_i^B,\,
D_i^A,\,
D_i^B,\,
y_i^{\mathrm{VQ}},\,
y_i^{\mathrm{MQ}},\,
y_i^{\mathrm{TA}},\,
y_i^{\mathrm{overall}}
\right).
\end{equation}
After filtering, the dataset contains 22,640 video pairs.

To validate pipeline quality, we conduct an expert spot-check with six domain professionals. Each professional independently annotates the visible artifacts of 50 randomly sampled videos, of which 20 videos are shared by all six annotators to measure inter-annotator agreement. Comparing the expert annotations against the automated pipeline output, the pipeline achieves a recall of 87.84\% and a precision of 84.37\%. On the 20 shared videos, we measure inter-annotator agreement with Fleiss' kappa: for each video, the artifacts reported by all annotators are pooled and semantically deduplicated, each pooled artifact is treated as an item, and each annotator's judgment on whether the artifact is present is treated as a binary rating. The resulting Fleiss' kappa across the six annotators is 0.78, indicating substantial agreement among human experts and supporting the reliability of the spot-check.

\subsection{Two-Stage Artifact-Aware Training}
\label{sec:training}
\begin{figure*}[t]
    \centering
    \includegraphics[width=\textwidth]{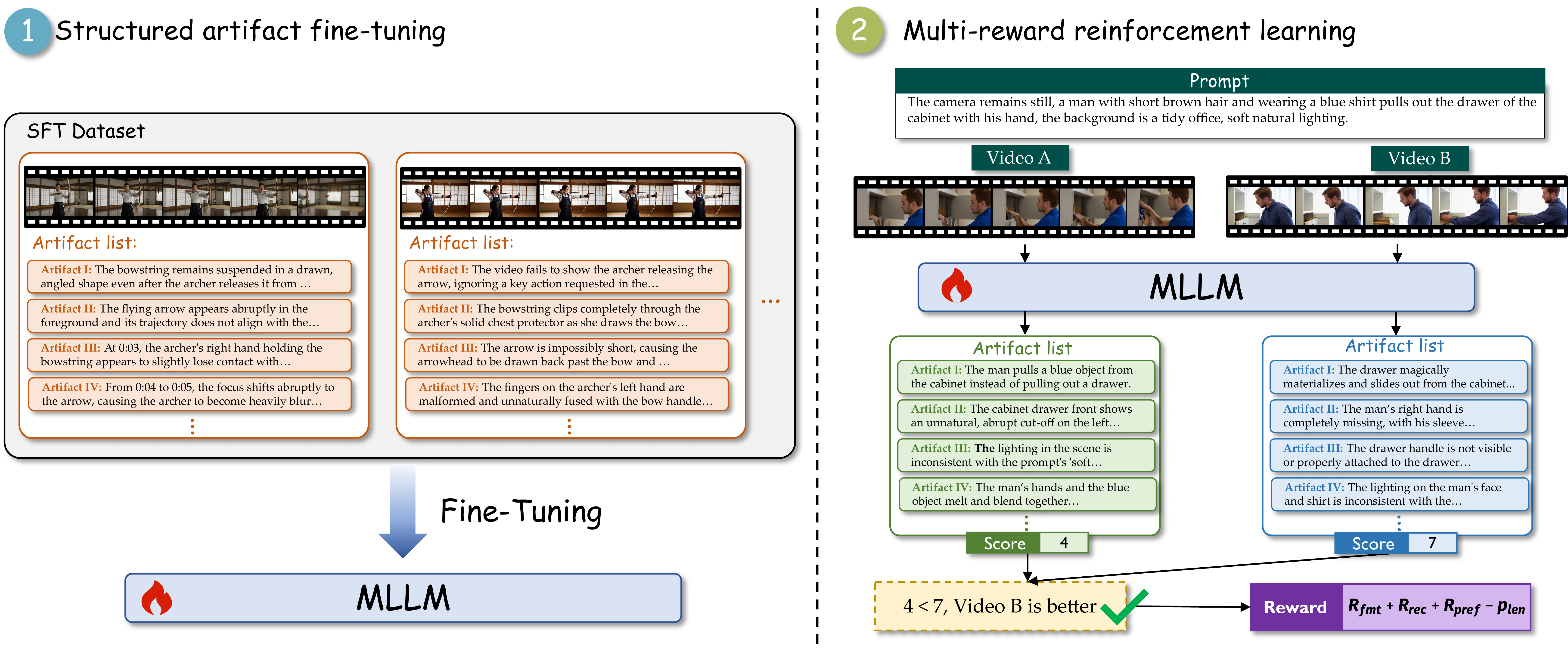}
    \caption{Overview of the two-stage training pipeline of ArtiVidRM.}
    \label{fig:train}
\end{figure*}


\paragraph{Stage I: Structured artifact fine-tuning.}
As shown in Figure~\ref{fig:train}, this stage establishes structured artifact identification through supervised fine-tuning (SFT). Given a text prompt $p$ and a generated video $v$, the model is trained to produce a structured list of visible generation artifacts. Each artifact contains a textual description and a severity weight. We decompose each video pair into two single-video examples and use the corresponding reference artifact list as the target response.

Let $z=(z_1,\ldots,z_T)$ denote the serialized structured annotation.
The model is optimized using the standard autoregressive objective:
\begin{equation}
\mathcal{L}_{\mathrm{SFT}}
=
-\sum_{t=1}^{T}
\log\pi_{\theta}(z_t\mid p,v,z_{<t}).
\label{eq:sft_objective}
\end{equation}

The resulting model, denoted by $\pi_{\mathrm{SFT}}$, is used to initialize
the reinforcement learning stage.

\paragraph{Stage II: Multi-reward reinforcement learning.}

This stage jointly aligns artifact identification and video preference
judgment through reinforcement learning.
The task is formulated as a two-turn interaction: in the first turn the
model observes Video~A and outputs a structured artifact list together with
an overall quality score; in the second turn it observes Video~B and
produces the same format.

We optimize the policy using Group Relative Policy Optimization
(GRPO)~\cite{shao2024deepseekmathpushinglimitsmathematical}. For each training instance, $G$ responses are
sampled from the current policy. Given their trajectory rewards
$\{R_i\}_{i=1}^{G}$, the normalized advantage is computed as
\begin{equation}
\hat{A}_i=
\frac{R_i-\bar{R}}{\sigma_R},
\label{eq:grpo_adv}
\end{equation}
where $\bar{R}$ and $\sigma_R$ denote the group mean and standard
deviation of rewards.

The policy is optimized using the standard GRPO objective with KL
regularization:
\begin{equation}
\mathcal{L}
=
\mathcal{L}_{\mathrm{GRPO}}
+
\beta D_{\mathrm{KL}}
(\pi_\theta\|\pi_{\mathrm{SFT}}),
\label{eq:grpo_loss}
\end{equation}
where $\pi_{\mathrm{SFT}}$ is the Stage~I checkpoint used as the
reference policy.

The trajectory reward is defined as
\begin{equation}
R=
\lambda_f r_{\mathrm{fmt}}
+\lambda_r r_{\mathrm{rec}}
+\lambda_p r_{\mathrm{pref}}
-\lambda_l p_{\mathrm{len}},
\label{eq:overall_reward}
\end{equation}
where the four terms evaluate output validity, artifact coverage,
preference correctness, and artifact-list length, respectively.

\paragraph{Format reward.}
The format reward awards partial credit per turn:
\begin{equation}
r_{\mathrm{fmt}}=\tfrac{1}{2}(q_1+q_2),
\label{eq:format_reward}
\end{equation}
where $q_t\in\{0,1\}$ indicates whether turn $t$ produces a parseable
artifact list and a valid quality score. Assigning 0.5 per turn preserves
a non-zero format signal even when only one turn fails, avoiding a hard
floor that discards partially valid trajectories.

\paragraph{Semantic artifact-recall reward.}
For each parseable turn, the predicted artifact list is compared with
the reference list using an LLM-based semantic matcher. A reference
artifact is considered recalled if at least one predicted artifact is
judged semantically equivalent. To avoid rewarding redundant
predictions, multiple predicted artifacts matching the same reference
artifact are counted only once. The recall reward is defined as the
fraction of reference artifacts that are successfully covered and is
averaged across the two turns. Following common practice, recall is set
to one when the reference artifact list is empty.

\paragraph{Preference reward.}
After both turns, we compare the predicted quality scores $s^A$ and
$s^B$. The preference is determined by the score difference using a tie
threshold $\tau$. A reward of 1 is assigned when the predicted
preference agrees with the human annotation, and 0 otherwise. This
reward is computed only when both turns produce valid quality scores;
otherwise, it is set to zero.

\begin{table*}[t]
\centering
\small
\renewcommand{\arraystretch}{1.15}%
\resizebox{\textwidth}{!}{%
\begin{tabular}{l cc cc cc}
\toprule
\multirow{2}{*}{\textbf{Model}}
  & \multicolumn{2}{c}{\textbf{GenAI-Bench}} & \multicolumn{2}{c}{\textbf{VBench-2.0}} & \multicolumn{2}{c}{\textbf{MonetBench}}  \\
\cmidrule(lr){2-3}\cmidrule(lr){4-5}\cmidrule(lr){6-7}
  & w/T & w/oT & w/T & w/oT & w/T & w/oT  \\
\midrule
\multicolumn{7}{c}{\textit{Discriminative Video Reward Models}} \\
\midrule
VideoScore~\cite{he2024videoscorebuildingautomaticmetrics}       & 0.481 & 0.719 & 0.429 & 0.459 & 0.491$^{*}$ & 0.549$^{*}$  \\
VideoAlign~\cite{liu2025improvingvideogenerationhuman}       & \underline{0.494} & 0.726 & 0.525 & \underline{0.473} & 0.505 & 0.594  \\
\midrule
\multicolumn{7}{c}{\textit{Generative Video Reward Models}} \\
\midrule
VisionReward~\cite{xu2026visionrewardfinegrainedmultidimensionalhuman}    & \textbf{0.518}$^{*}$ & \underline{0.744$^{*}$} & -- & -- & \textbf{0.640$^{*}$} & \textbf{0.721$^{*}$}  \\
UnifiedReward~\cite{wang2026unifiedrewardmodelmultimodal}   & 0.459 & 0.689 & \underline{0.541} & 0.402 & 0.434 & 0.404  \\
VideoScore2~\cite{he2025videoscore2thinkscoregenerative}      & 0.456 & 0.678 & 0.404 & 0.468 & 0.449 & 0.453 \\
\midrule
\multicolumn{7}{c}{\textit{Our Models}} \\
\midrule
\textbf{ArtiVidRM(Ours)}
                 & \textbf{0.518} & \textbf{0.753} & \textbf{0.572} & \textbf{0.524} & \underline{0.606} & \underline{0.697} \\
\bottomrule
\end{tabular}%
}
\caption{Pairwise preference prediction accuracy on out-of-domain benchmarks.
\textbf{w/T}: accuracy with ground-truth ties retained;
\textbf{w/oT}: accuracy after removing ground-truth tie pairs.
Each model uses its default tie threshold when applicable.
Best results are shown in \textbf{bold} and second-best results are \underline{underlined}; ``--'' indicates unavailable results. $^{*}$ denotes results directly taken from the corresponding original publications.}
\label{tab:ood}
\end{table*}

\paragraph{Asymmetric length penalty.}
To discourage over-generation not captured by recall, we apply an
asymmetric length penalty. Let $\hat{n}$ and $n$ denote the predicted
and reference artifact counts for a single video:
\begin{equation}
p_{\mathrm{len}}=\begin{cases}
\min\!\left(1,\;\dfrac{\hat{n}-n}{n}\right) & \hat{n}>n\\[6pt]
\dfrac{1}{2}\min\!\left(1,\;\dfrac{n-\hat{n}}{n}\right) & \hat{n}\leq n
\end{cases}
\label{eq:len_penalty}
\end{equation}
Over-generation is penalized at full rate; under-generation at half rate,
since missing artifacts are already penalized by the recall reward.
No penalty is applied when the reference list is empty.
The penalties from both turns are averaged to obtain $p_{\mathrm{len}}$
in Eq.~\eqref{eq:overall_reward}.

\section{Experiments}

\paragraph{Training Setting.}
We instantiate ArtiVidRM with Qwen3-VL-8B-Instruct as the backbone.
During Stage II, we optimize the model using GRPO with a rollout size of
8 for each video pair. The actor is trained with a learning rate of
$2.5\times10^{-7}$ and a maximum gradient norm of 1.0. The reference
policy is initialized from the Stage I checkpoint, with the KL
regularization coefficient set to $\beta=0.01$.

Unless otherwise specified, the reward weights are
$\lambda_f=0.15$, $\lambda_r=0.40$, $\lambda_p=0.45$, and
$\lambda_l=0.05$, and the tie threshold is $\tau=2$. All experiments
are conducted on a server with eight NVIDIA H100 GPUs using the
\texttt{verl} framework and vLLM for rollout generation.

\subsection{ArtiVidRM for Video Preference Prediction}
Video preference prediction is evaluated from three complementary
aspects: pairwise preference prediction, multi-video ranking, and
pointwise quality scoring. All evaluations are conducted on out-of-domain
benchmarks with no overlap with the training data.


\paragraph{Pairwise video preference prediction.}
To evaluate the generalization capability of ArtiVidRM, we conduct
experiments on three out-of-domain benchmarks: GenAI-Bench~\cite{li2024genaibenchevaluatingimprovingcompositional},
VBench-2.0~\cite{zheng2025vbench20advancingvideogeneration}, and MonetBench~\cite{xu2026visionrewardfinegrainedmultidimensionalhuman}.

Table~\ref{tab:ood} summarizes the pairwise preference prediction
results on the three benchmarks. On GenAI-Bench, it achieves
the best performance under both settings, tying with VisionReward under
the w/T setting. On VBench-2.0, ArtiVidRM achieves the best performance
under both the w/T and w/oT settings. On MonetBench, it ranks second
only to VisionReward.

\begin{figure}[t]
    \centering
    \begin{subfigure}[c]{0.51\linewidth}
        \centering
        \includegraphics[width=\linewidth]{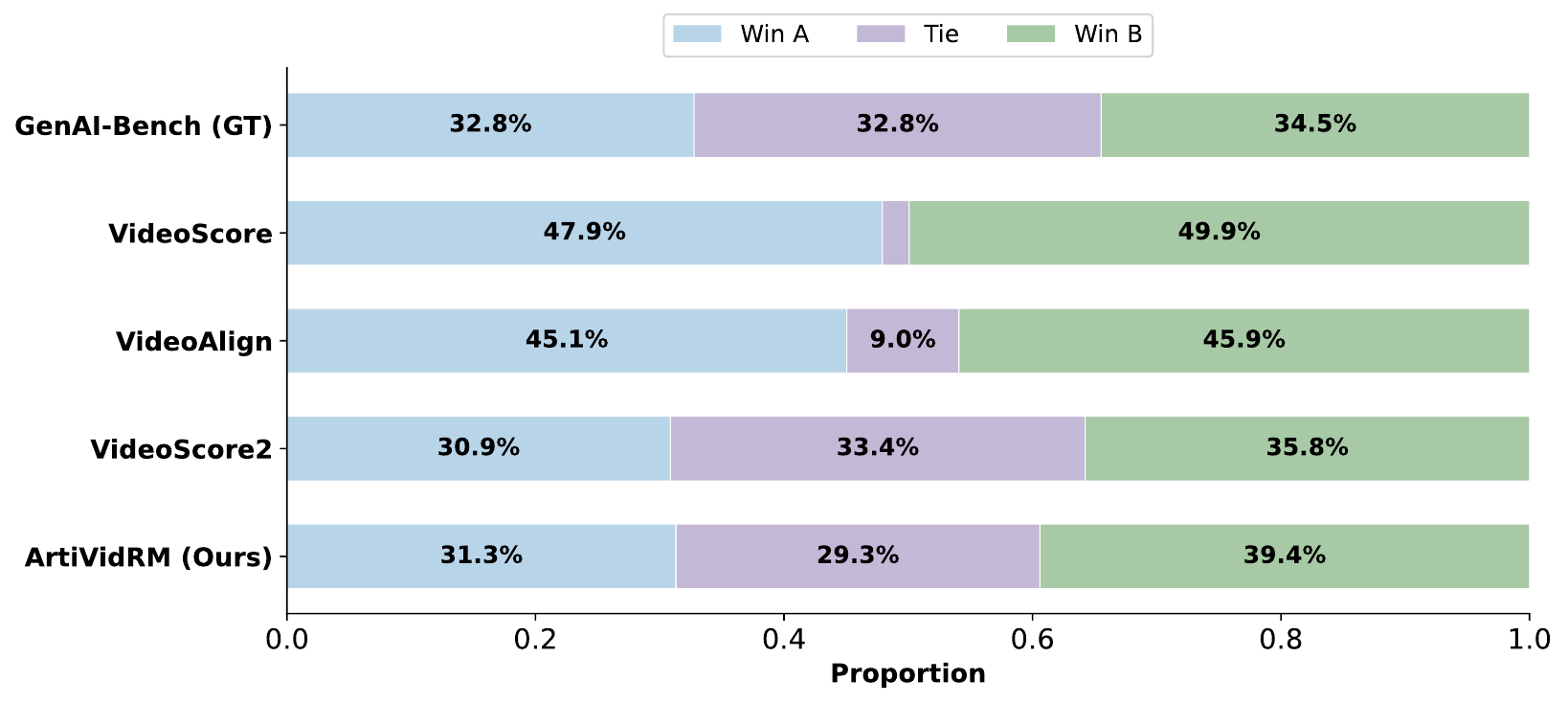}
        \subcaption{}
        \label{fig:placeholder}
    \end{subfigure}\hfill
    \begin{subfigure}[c]{0.47\linewidth}
        \centering
        \includegraphics[width=\linewidth]{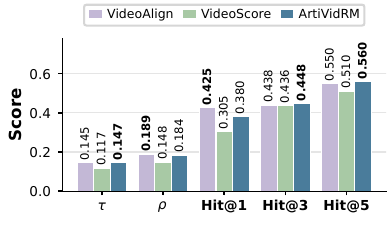}
        \subcaption{}
        \label{fig:ranking}
    \end{subfigure}
    \caption{(a) Preference prediction distributions of different reward models on GenAI-Bench.
    (b) Multi-video ranking performance on T2VQA-DB.}
\end{figure}

Beyond the overall preference prediction accuracy, we further analyze
the preference distributions produced by different reward models on
GenAI-Bench.
As shown in Figure~\ref{fig:placeholder}, ArtiVidRM produces a
preference distribution that is closer to the ground-truth
annotations, particularly in the proportion of tie predictions. In
contrast, VideoScore~\cite{he2024videoscorebuildingautomaticmetrics} and VideoAlign~\cite{liu2025improvingvideogenerationhuman} predict ties much less frequently,
showing a stronger tendency toward binary preferences.


\paragraph{Multi-video Ranking.}
We evaluate the ability of ArtiVidRM to rank multiple candidate videos
using T2VQA-DB~\cite{kou2024subjectivealigneddatasetmetrictexttovideo}, which contains 1,000 text prompts, each
paired with 10 videos generated by different text-to-video models.
Each video is annotated with a Mean Opinion Score (MOS) collected from
27 human raters.

For each prompt, we independently score all candidate videos and rank
them according to the predicted quality scores. We then evaluate the
predicted ordering using Kendall's $\tau$, Spearman's $\rho$, and
Hit@$K$, and report the average results over all prompts. Kendall's
$\tau$ and Spearman's $\rho$ measure the agreement between the
predicted ordering and the human MOS ordering, while Hit@$K$ measures
the percentage of prompts for which the human top-ranked video is
retrieved within the model's top-$K$ predictions.

Figure~\ref{fig:ranking} shows that ArtiVidRM consistently performs
competitively across all ranking metrics. In particular, it achieves
the highest Kendall's $\tau$ together with the best Hit@3 and Hit@5,
indicating that the predicted scores preserve the overall ordering of
candidate videos while reliably identifying high-quality videos.
VideoAlign attains higher Hit@1 and Spearman's $\rho$.


\begin{figure}[t]
    \centering
    \begin{subfigure}[c]{0.515\linewidth}
        \centering
        \includegraphics[width=\linewidth]{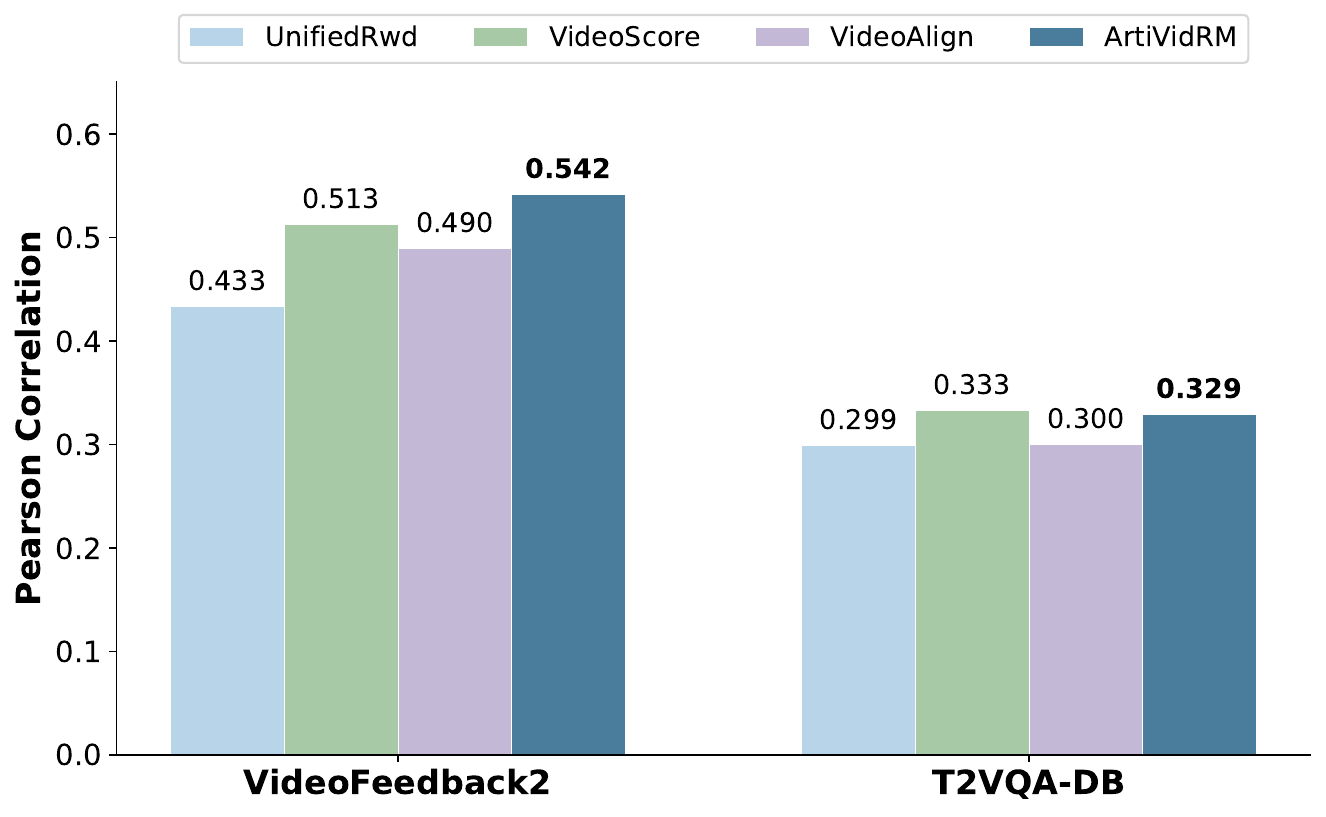}
        \subcaption{}
        \label{fig:Pearson}
    \end{subfigure}\hfill
    \begin{subfigure}[c]{0.465\linewidth}
        \centering
        \includegraphics[width=\linewidth]{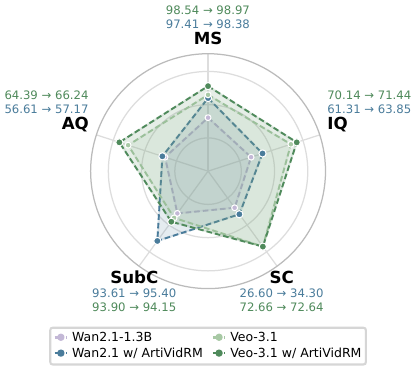}
        \subcaption{}
        \label{fig:vbench}
        \label{fig:vbench-veo}
    \end{subfigure}

    \vspace{4pt}
    \begin{subfigure}[b]{0.485\linewidth}
        \centering
        \includegraphics[width=\linewidth]{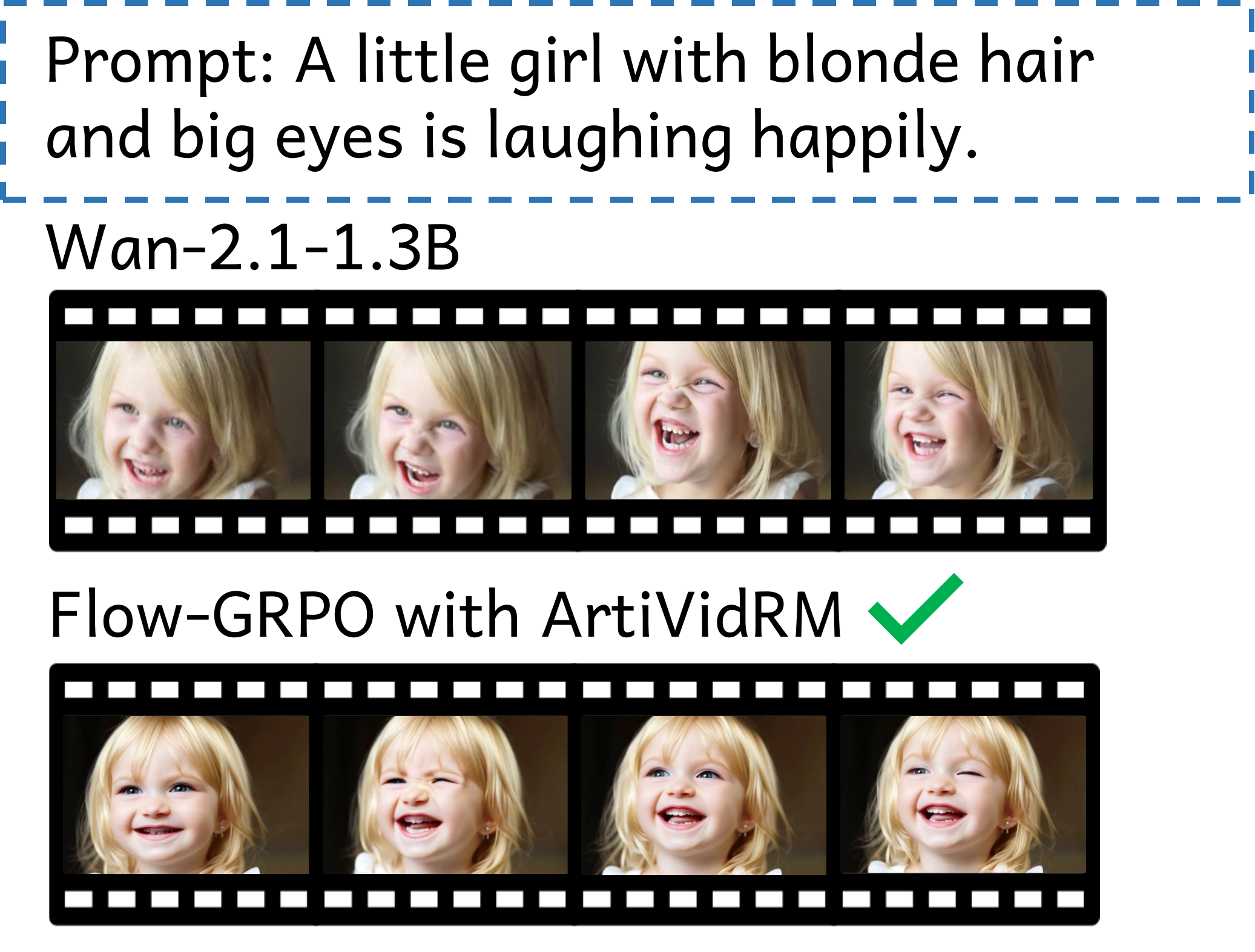}
        \subcaption{}
        \label{fig:human-wan}
    \end{subfigure}\hfill
    \begin{subfigure}[b]{0.485\linewidth}
        \centering
        \includegraphics[width=\linewidth]{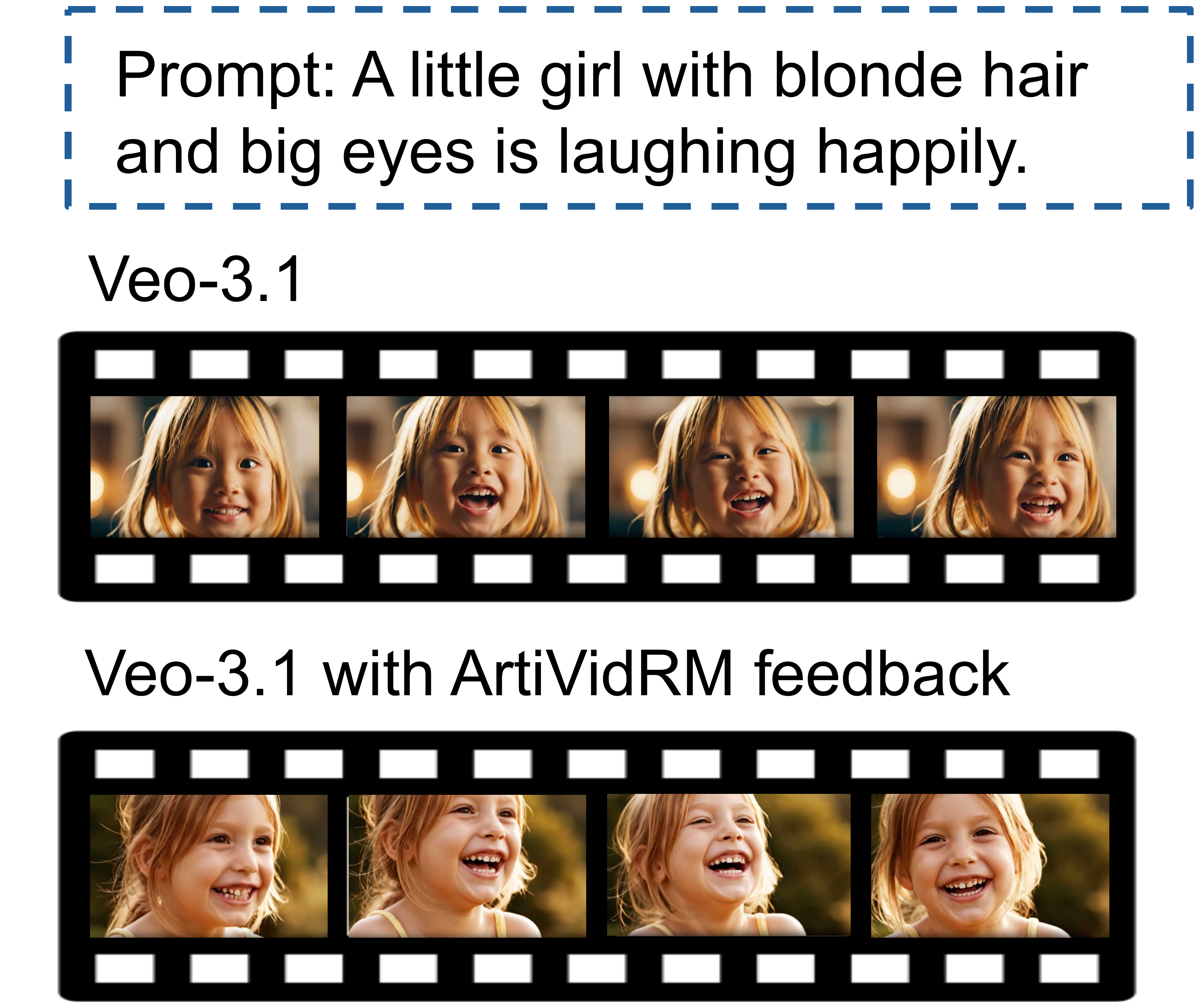}
        \subcaption{}
        \label{fig:human-veo}
    \end{subfigure}
    \caption{(a) Pointwise quality scoring performance measured by Pearson Correlation between predicted quality scores and human quality ratings.
    (b) VBench evaluation results of the diffusion model (Wan2.1-1.3B) before and after optimization with ArtiVidRM, and of Veo-3.1 before and after refinement with ArtiVidRM feedback.
    MS: Motion Smoothness; IQ: Imaging Quality; SC: Scene; SubC: Subject Consistency; AQ: Aesthetic Quality.
    Each axis is scaled to its own value range; the raw scores (before $\rightarrow$ after) are listed next to each axis.
    (c) A qualitative comparison between Wan-2.1 before and after optimization with ArtiVidRM.
    (d) A qualitative comparison between Veo-3.1 before and after refinement with ArtiVidRM feedback.}
\end{figure}

\paragraph{Scoring Consistency.}
Beyond pairwise preference prediction and multi-video ranking, we further
evaluate the consistency between predicted quality scores and human
quality judgments. Following prior work, we compute the Pearson
Correlation between the predicted quality scores and the corresponding
human quality ratings on two public benchmarks.

Figure~\ref{fig:Pearson} shows the pointwise quality scoring
results. ArtiVidRM achieves competitive Pearson Correlation across both
benchmarks, outperforming all baselines on VideoFeedback2 and matching
the performance of the strongest baseline on T2VQA-DB~\cite{kou2024subjectivealigneddatasetmetrictexttovideo}.


\begin{table*}[t]
\centering
\renewcommand{\arraystretch}{1.15}
\resizebox{\textwidth}{!}{%
\begin{tabular}{lccc cc cc cc cc}
\toprule
\multirow{2}{*}{\textbf{Variant}}
  & \multicolumn{3}{c}{\textbf{Components}}
  & \multicolumn{2}{c}{\textbf{GenAI-Bench}}
  & \multicolumn{2}{c}{\textbf{VBench-2.0}}
  & \multicolumn{2}{c}{\textbf{VisionRewardDB}}
  & \multicolumn{2}{c}{\textbf{VideoFeedback2}} \\
\cmidrule(lr){2-4}\cmidrule(lr){5-6}\cmidrule(lr){7-8}\cmidrule(lr){9-10}\cmidrule(l){11-12}
  & SFT & $r_{\mathrm{rec}}$ & $r_{\mathrm{len}}$
  & w/T & w/oT & w/T & w/oT & w/T & w/oT & w/T & w/oT \\
\midrule
Base (Qwen3-VL-8B)
  & \ding{55} & \ding{55} & \ding{55}
  & 0.435 & 0.497 & 0.508 & 0.226 & 0.346 & 0.411 & 0.543 & 0.575 \\
w/o SFT
  & \ding{55} & \ding{51} & \ding{51}
  & 0.342 & 0.431 & 0.580 & 0.183 & 0.232 & 0.330 & 0.473 & 0.483 \\
w/o Recall Reward
  & \ding{51} & \ding{55} & \ding{51}
  & 0.469 & 0.624 & 0.407 & 0.358 & 0.398 & 0.554 & 0.551 & 0.599 \\
w/o Length Penalty
  & \ding{51} & \ding{51} & \ding{55}
  & 0.459 & 0.605 & 0.392 & 0.312 & 0.439 & 0.536 & 0.574 & 0.622 \\
\midrule
\textbf{ArtiVidRM }
  & \ding{51} & \ding{51} & \ding{51}
  & 0.518 & 0.753 & 0.572 & 0.524 & 0.606 & 0.697 & 0.567 & 0.619  \\
\bottomrule
\end{tabular}%
}
\caption{Ablation study on training components.
SFT = supervised fine-tuning stage;
$r_{\mathrm{rec}}$ = defect recall reward;
$r_{\mathrm{len}}$ = length penalty.
w/T and w/oT metrics are consistent with Table~\ref{tab:ood}.}
\label{tab:ablation}
\end{table*}

\subsection{ArtiVidRM for Video Diffusion Optimization}

To evaluate the practical utility of ArtiVidRM, we use it as the reward
model for reinforcement learning fine-tuning of a text-to-video
diffusion model. Specifically, we fine-tune Wan2.1-1.3B~\cite{wan2025wanopenadvancedlargescale}
with Flow-GRPO~\cite{liu2025flowgrpotrainingflowmatching}, using ArtiVidRM to provide per-sample
rewards during training. We evaluate the resulting model against the
original Wan2.1-1.3B on the VBench~\cite{huang2023vbenchcomprehensivebenchmarksuite} benchmark.

\paragraph{Quantitative results.}
Figure~\ref{fig:vbench} reports scores on five VBench dimensions.
Compared with the base model, the ArtiVidRM-trained model
improves Motion Smoothness from 97.41\% to 98.38\% and Imaging Quality
from 61.31\% to 63.85\%, demonstrating that the reward signal successfully
propagates defect-level feedback into the generative model.

\paragraph{Human preference evaluation.}
To further assess the perceptual quality of the optimized video
generation model, we conduct a human evaluation. We randomly sample 100
prompts from GenAI-Bench and generate videos using both the original
and the ArtiVidRM-optimized models. Two expert annotators each evaluate
50 video pairs in a blind pairwise comparison, selecting the preferred
video or marking a tie when the quality difference is negligible.
Overall, the videos generated by the ArtiVidRM-optimized model are
preferred in \textbf{60\%} of the comparisons, while \textbf{14\%} are judged as ties.
Figure~\ref{fig:human-wan} shows a representative example where the
ArtiVidRM-optimized model mitigates the distortion artifacts in the
original generation and improves the overall visual quality, and additional examples are provided in the appendix.

\subsection{ArtiVidRM for Closed-Source Video Refinement}

Beyond training open-weight diffusion models, ArtiVidRM can also be
used to refine closed-source video generators without access to model
weights. Given an initial generated video, ArtiVidRM predicts a
structured artifact list describing its defects. The artifact list is
converted into natural language feedback and appended to the original
prompt, guiding the generator to produce a refined video in a second
generation round. We evaluate the refined videos using both VBench and
human preference judgments.

\paragraph{Quantitative results.}
As shown in Figure~\ref{fig:vbench-veo}, ArtiVidRM feedback improves the quality of Veo-3.1 generations, yielding higher scores on four of the five VBench dimensions. The gains concentrate on the appearance-related dimensions, with Aesthetic Quality improving by 1.85 and Imaging Quality by 1.30, whereas Scene and Subject Consistency remain nearly unchanged. This pattern matches the nature of the feedback: the predicted artifacts mostly describe rendering, lighting, and texture defects, which the generator can address in a second round, while the scene layout is constrained by the original prompt that is kept unchanged. The improvements are smaller than those obtained by reward-guided fine-tuning of Wan2.1-1.3B, which is expected because the generator weights are not updated and the feedback enters only through the prompt.

\paragraph{Human preference evaluation.}
We conduct the same human evaluation for the closed-source Veo-3.1. The ArtiVidRM-refined videos are preferred in \textbf{71\% }of the comparisons, with \textbf{18\%} judged as ties. In the remaining cases the original video is preferred, which we attribute mainly to sampling variance: the refined video is a fresh generation rather than an edit of the original, so a second round can introduce new artifacts even when the reported ones are resolved. More detailed results are provided in the appendix.

\subsection{Ablation Study}

We conduct ablation studies to evaluate the contribution of each
training component. Specifically, we compare five variants:
Base, which directly uses the pretrained Qwen3-VL-8B-Instruct;
w/o SFT, which removes supervised fine-tuning;
w/o Recall Reward, which removes the semantic recall reward;
w/o Length Penalty, which removes the length penalty; and the
full ArtiVidRM. Unless otherwise specified, all variants use
the same model architecture, training settings, and evaluation
protocol.

Table~\ref{tab:ablation} summarizes the ablation results. Compared with the pretrained Qwen3-VL-8B, the full ArtiVidRM achieves the best performance on most benchmarks. Removing the supervised fine-tuning stage causes a noticeable performance drop, highlighting the importance of structured artifact learning before reinforcement learning.

The semantic recall reward also plays an important role. Removing
$r_{\mathrm{rec}}$ consistently degrades preference prediction
performance, indicating that encouraging comprehensive artifact
identification enables the model to capture more complete quality
signals, which in turn benefits downstream reward estimation.

We further compare \textbf{the average length of the predicted artifact lists} across all benchmarks. The average length is 1.7 for Base, 5.9 for w/o SFT, 4.7 for the full model, 4.0 for w/o Recall Reward, and 2.7 for w/o Length Penalty. These results show that the semantic recall reward encourages more comprehensive artifact prediction, the length penalty encourages active artifact generation, and SFT stabilizes artifact prediction. More detailed results are provided in the appendix.

\section{Conclusion}

We present ArtiVidRM, an artifact-as-rubric reward model for video
generation. By explicitly modeling structured artifacts, ArtiVidRM
achieves strong performance on video preference prediction and enables
both reward-guided optimization of open-weight diffusion models and
feedback-based refinement of closed-source video generators. We hope
this work inspires future research on video reward modeling.

\bibliography{references}
\bibliographystyle{arxiv}

\clearpage
\appendix

\begin{figure*}[t]
    \centering
    \includegraphics[width=\textwidth]{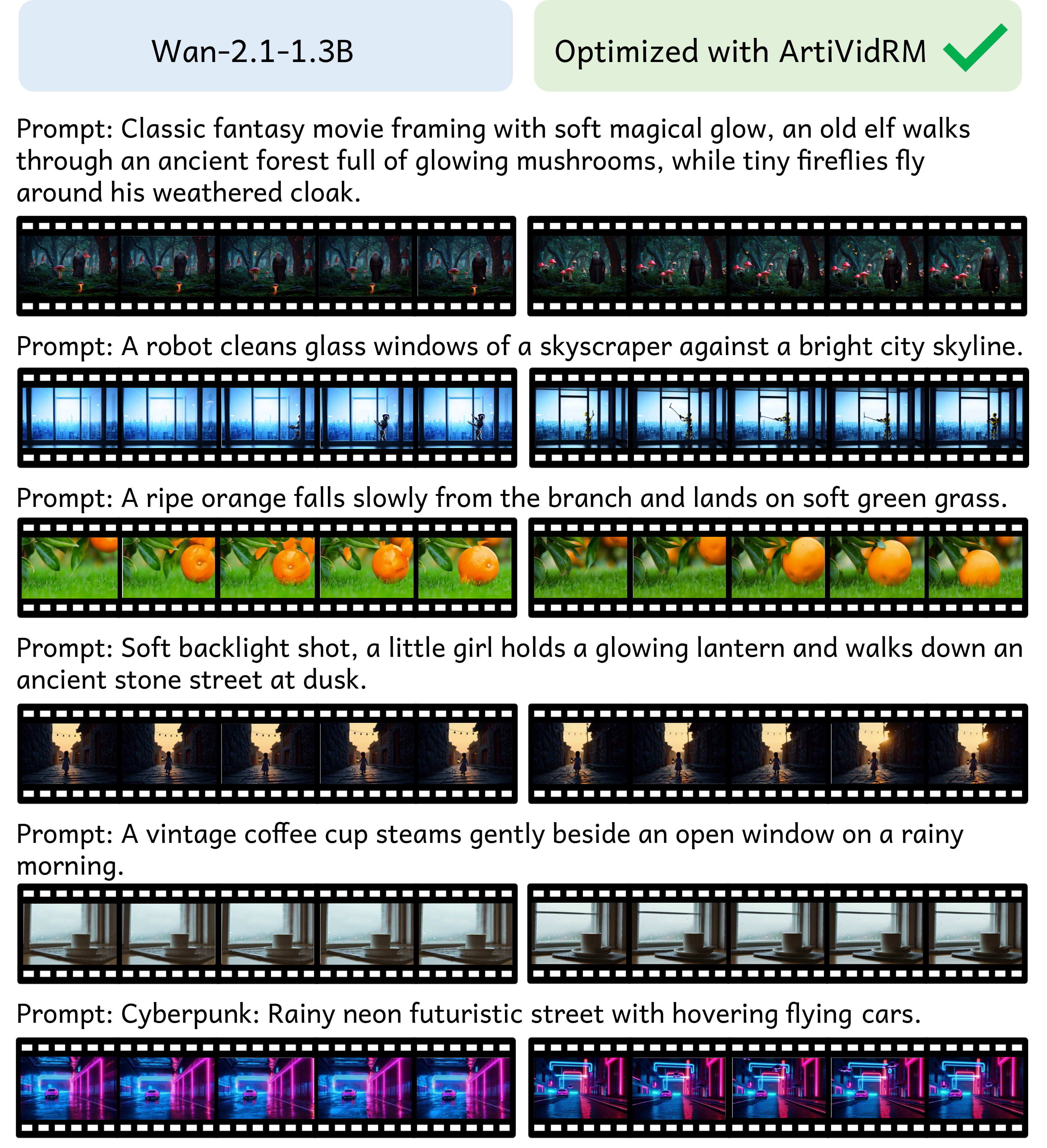}
    \caption{Representative qualitative comparisons of Wan-2.1-1.3B before and after optimization with ArtiVidRM.}
    \label{fig:wan-cases}
\end{figure*}

\section{Training Prompt}

We use the following two-turn prompt during training.

\begin{tcolorbox}[
    colback=gray!5,
    colframe=gray!50,
    boxrule=0.5pt,
    arc=2pt,
    left=6pt,
    right=6pt,
    top=6pt,
    bottom=6pt,
    breakable
]
\textbf{System Prompt}

\smallskip

You are a video defect detection expert. Given a text prompt and an
AI-generated video, identify all visible defects in the video and assign
an overall quality score (1 = worst, 10 = best).

For each defect, specify:

\begin{itemize}
    \item \texttt{"issue"} --- one concise sentence describing a specific,
    observable defect;
    \item \texttt{"weight"} --- severity: 1 (minor), 2 (moderate), or 3
    (severe).
\end{itemize}

Only report defects that are clearly visible and unambiguous. If no defects
are visible, output an empty defect list.

Output only valid JSON in exactly the following format:

\begin{verbatim}
{"defects": [{"issue": "...", "weight": <1|2|3>}, ...],
 "score": <1-10 integer>}
\end{verbatim}

\medskip
\textbf{Turn 1: Video A}

\smallskip

\textbf{User}

\begin{verbatim}
Text prompt: "{text_prompt}"

Video A:
<video>

Evaluate Video A. Identify all visible defects and assign an
overall quality score.

{"defects": [{"issue": "...", "weight": <1|2|3>}, ...],
 "score": <1-10 integer>}
\end{verbatim}

\medskip
\textbf{Turn 2: Video B}

\smallskip

\textbf{User}

\begin{verbatim}
---VIDEO_B---
Text prompt: "{text_prompt}"

Evaluate Video B. Identify all visible defects and assign an
overall quality score.

{"defects": [{"issue": "...", "weight": <1|2|3>}, ...],
 "score": <1-10 integer>}
\end{verbatim}
\end{tcolorbox}

\section{Cases of Wan}

Figure~\ref{fig:wan-cases} presents additional qualitative
examples before and after optimization with ArtiVidRM. Across diverse
prompts, the optimized model consistently produces videos with fewer
visible artifacts and improved visual quality, further supporting the
quantitative results reported in the main paper.


\section{Cases of Veo}

\begin{figure*}[t]
    \centering
    \includegraphics[width=\textwidth]{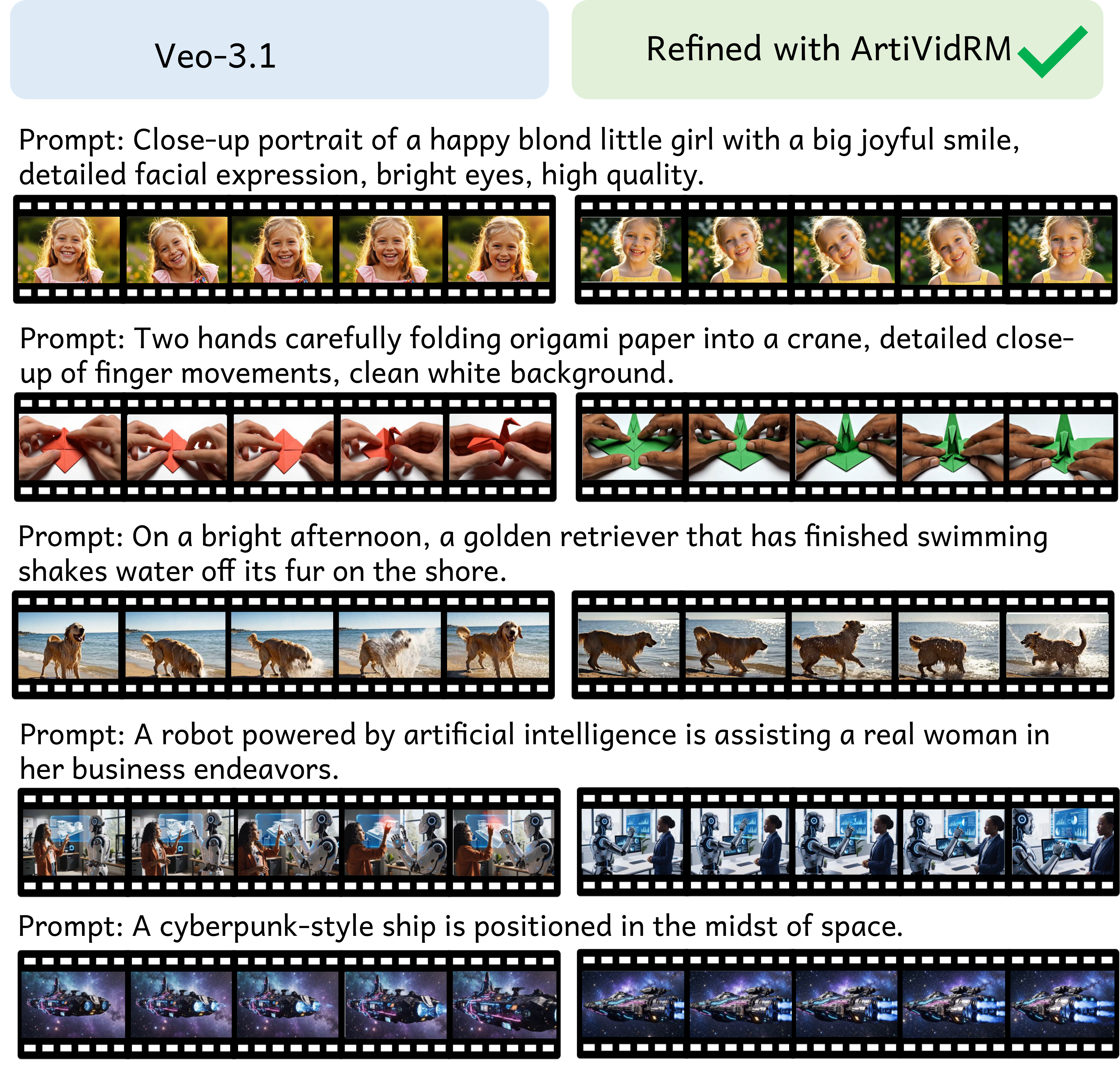}
    \caption{Additional qualitative comparisons between the original Veo-3.1 and the videos refined with ArtiVidRM across diverse prompts.}
    \label{fig:veo-cases}
\end{figure*}

Figure~\ref{fig:veo-cases} provides additional qualitative
examples of the closed-source Veo-3.1 before and after refinement with
ArtiVidRM feedback. The refined videos consistently exhibit fewer
visible artifacts and improved visual quality across diverse prompts,
further supporting the results reported in the main paper.

\section{Analysis of Artifact List Quality in the Ablation Study}

\begin{figure*}[t]
    \centering
    \includegraphics[width=\textwidth]{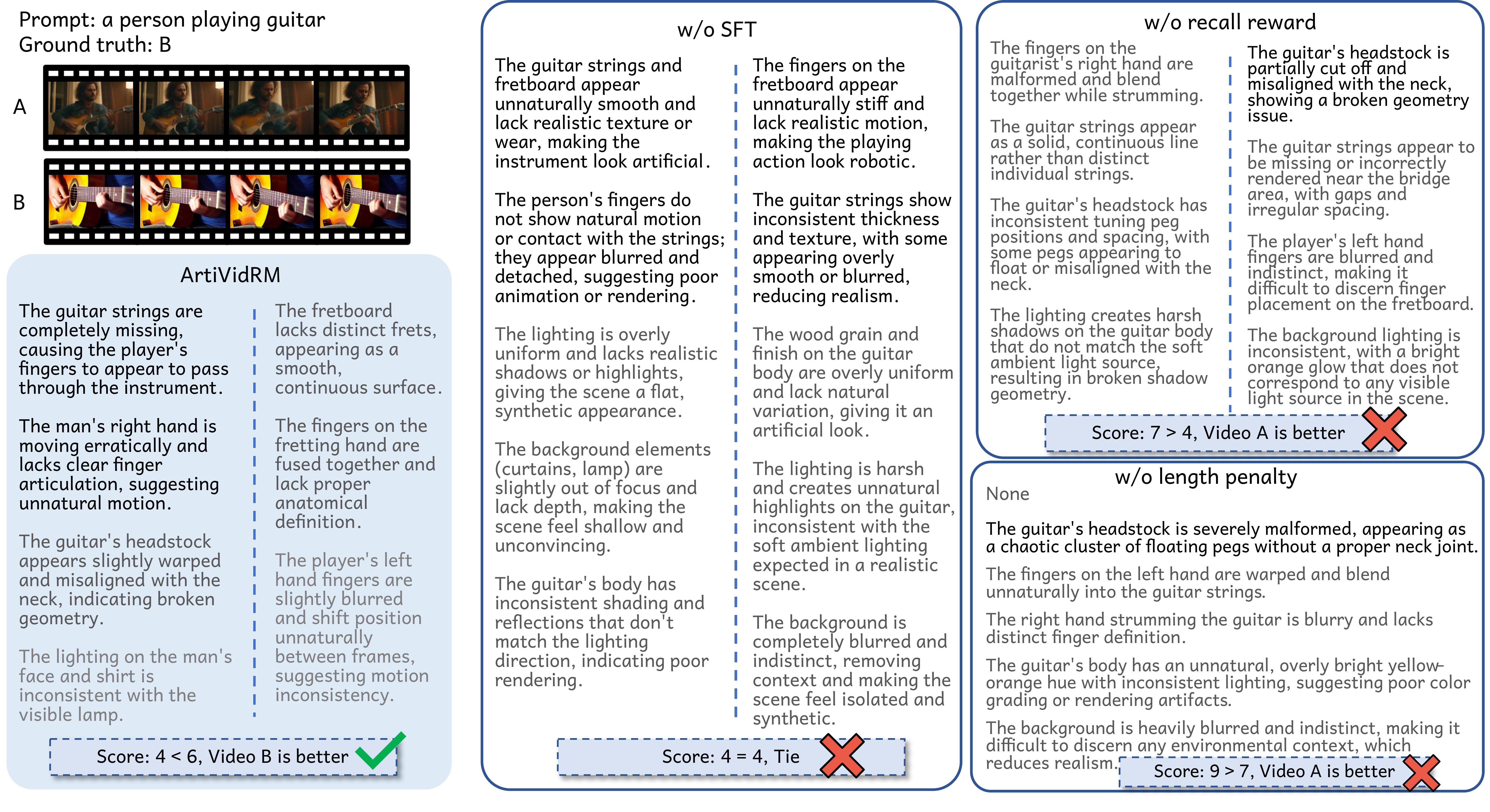}
    \caption{Representative artifact lists generated by different ablation
variants. Darker text indicates higher artifact importance predicted by
the model.}
    \label{fig:ab-list}
\end{figure*}

Figure~\ref{fig:ab-list} provides a representative comparison of
the artifact lists generated by different ablation variants. Darker text
indicates artifacts assigned higher importance by the model. Compared
with the full ArtiVidRM, removing the supervised fine-tuning stage
results in verbose and unstable artifact descriptions, many of which are
redundant or weakly related to the final preference prediction.
Removing the semantic recall reward causes the model to overlook several
important defects, leading to less comprehensive artifact coverage.
Without the length penalty, the model becomes overly conservative and
focuses only on a few prominent artifacts while omitting many secondary
defects. In contrast, the full model produces a balanced artifact list
that is both comprehensive and focused, assigning higher weights to the
most preference-relevant artifacts and ultimately leading to the correct
pairwise decision.

\end{document}